\documentclass[
]{ceurart}

\usepackage{minted}

\setminted{breaklines=true}

\hypersetup{breaklinks=true}
\usepackage{graphicx}
\usepackage{tikz}
\usepackage{hyperref}
\usepackage{amsmath}
\usepackage{enumitem}
\usepackage{subcaption}
\usepackage{caption}
\usepackage{minted}
\usetikzlibrary{shapes.misc, positioning,fit}
\setminted{breaklines=true}

\begin{document}

\copyrightyear{2022}
\copyrightclause{Copyright for this paper by its authors.
  Use permitted under Creative Commons License Attribution 4.0
  International (CC BY 4.0).}

\conference{In: R. Campos, A. Jorge, A. Jatowt, S. Bhatia, M. Litvak (eds.): Proceedings of the Text2Story'22 Workshop, Stavanger (Norway), 10-April-2022}

\title{MARCUS: An Event-Centric NLP Pipeline that generates Character Arcs from Narratives}

\author[1]{Sriharsh Bhyravajjula}[%
email=s.bhyravajjula@research.iiit.ac.in,
]
\author[1]{Ujwal Narayan}[%
email=ujwal.narayan@research.iiit.ac.in,
]
\author[1]{Manish Shrivastava}[%
email=m.shrivastava@iiit.ac.in,
]
\address[1]{International Institute of Information Technology, Hyderabad}

\begin{abstract}
Character arcs are important theoretical devices employed in literary studies to understand character journeys, identify tropes across literary genres, and establish similarities between narratives. This work addresses the novel task of computationally generating event-centric, relation-based character arcs from narratives. Providing a quantitative representation for arcs brings tangibility to a theoretical concept and paves the way for subsequent applications. We present MARCUS (Modelling Arcs for Understanding Stories), an NLP pipeline that extracts events, participant characters, implied emotion, and sentiment to model inter-character relations. MARCUS tracks and aggregates these relations across the narrative to generate character arcs as graphical plots. We generate character arcs from two extended fantasy series, Harry Potter and Lord of the Rings. We evaluate our approach before outlining existing challenges, suggesting applications of our pipeline, and discussing future work.
\end{abstract}

\begin{keywords}
  Computational Literary Studies \sep
  Character Arcs \sep
  Relations \sep
  Events \sep
  Narratives
\end{keywords}

\maketitle

\section{Introduction}
Characters in narratives attempt to influence their circumstances to resolve conflict, while circumstance itself shapes the characters with events that develop them \cite{weiland}. This character journey is integral to narratives \cite{vonnegutshapes}; good stories are driven by the transformative journeys of compelling characters. This work addresses the challenge of quantifying these journeys using character arcs modeled around events and relations.

The field of computational literary studies aims to understand, represent and generate narratives. Existing research has focused on various perspectives, notably plot units \cite{10.5555/2380816.2380893, Lehnert1981PlotUA}, social network extraction from narratives \cite{agarwal-etal-2013-sinnet}, and character-centric approaches 
\cite{inproceedings, bamman-etal-2014-bayesian}. This work builds upon the latter to extract characters from narratives using co-reference resolution and clustering techniques. We find inspiration in research exploring events in narratives \cite{sims-etal-2019-literary} and argue that the transformation of characters is a consequence of events that involve these characters. We also take advantage of recent advances in semantic role labeling \cite{Shi2019SimpleBM} to understand the function of characters in events and build on advancements in emotion analysis \cite{Demszky2020GoEmotionsAD, zad-finlayson-2020-systematic} to represent how events affect both agent and recipient characters quantitatively. 

\begin{figure}[h]
  \centering
  \includegraphics[width=3.6in]{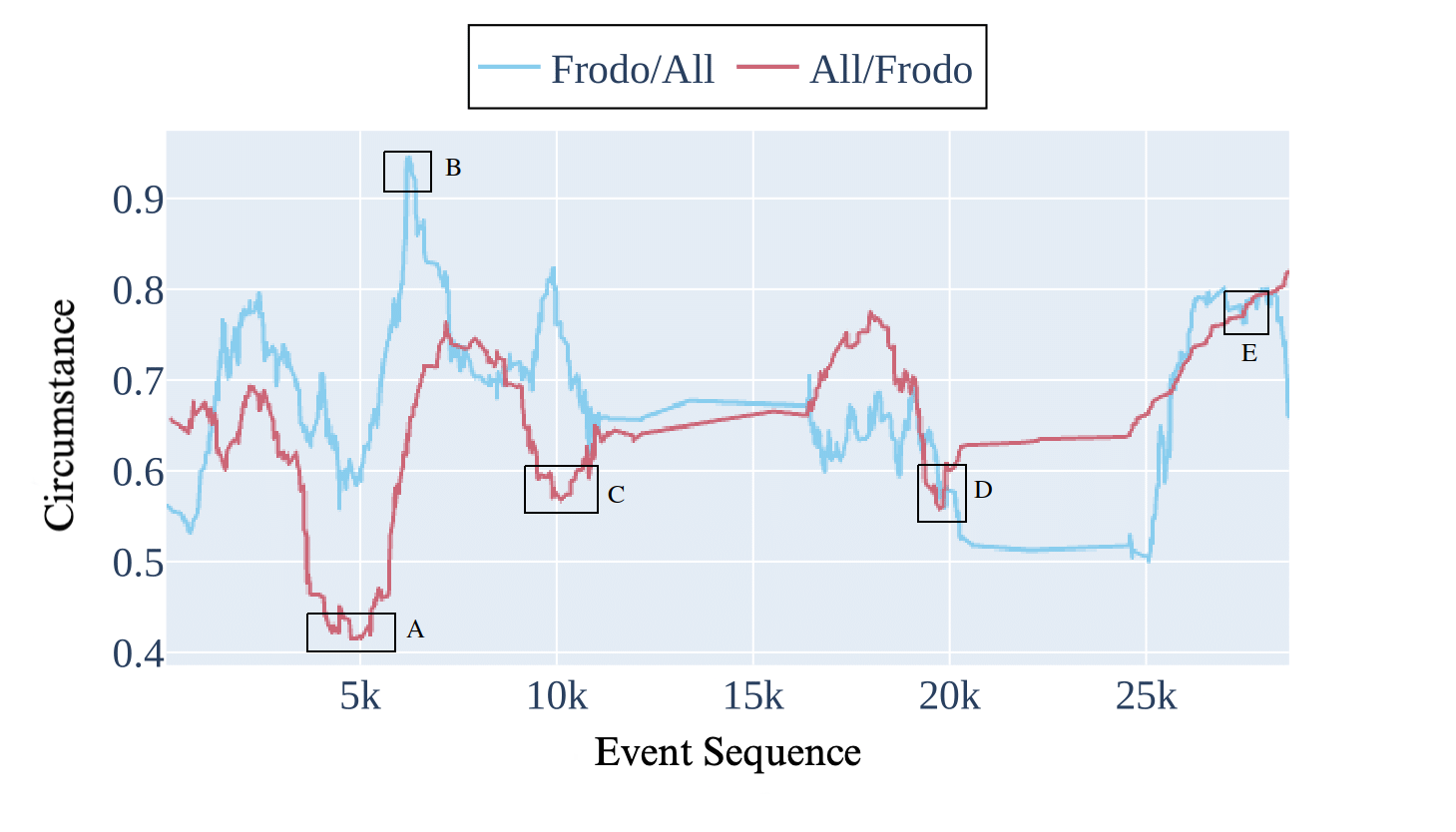}
  \caption{Relation-Based Character Arc for Frodo Baggins across The Lord of the Rings trilogy. The blue line represents Frodo the \textit{actor}, and the red line represents Frodo the \textit{experiencer}. \textbf{A}: Frodo is stabbed by the poisoned blade of the Nazgul at Weathertop (valley). \textbf{B}: Frodo reunites with loved ones at Rivendell (peak). \textbf{C}: Frodo is in grief after the wizard Gandalf falls to a Balrog (valley). \textbf{D}: Frodo is attacked by Shelob in her lair (valley). \textbf{E}: Frodo succeeds in his quest and returns home to The Shire (peak).}
  \label{fig:introduction}
\end{figure}

Our approach differs from recent work in the closely related field of extracting relationships between characters in terms of goals and linguistic features. Some existing approaches either lack the nuance of emotions \cite{DBLP:journals/corr/SrivastavaCM15, DBLP:journals/corr/ChaturvediSDD15} or do not leverage semantic role labels for characters \cite{iyyer-etal-2016-feuding}. Other approaches fail to utilize fine-grained sentiments \cite{Rashid2018CharacterizingIA} or do not share our event-based perspective on relationships \cite{DBLP:journals/corr/abs-1903-12453, 10.5555/3298023.3298028}. While there has been progress in the field of emotion arcs, it has been centered around plot rather than character, capturing a shift in sentiment rather than a shift in circumstance \cite{somasundaran-etal-2020-emotion, DBLP:journals/corr/ReaganMKDD16}. To the best of our knowledge, we believe that our task and approach are novel additions to the field. 

\section{Key Concepts}
\label{sec:concepts}
\subsection{Events}
We borrow from \cite{sims-etal-2019-literary} and focus solely on \emph{events} with asserted \emph{realis} (depicted as actually taking place, with specific participants at a specific time) instead of those with other epistemic modalities (hypotheticals, future events). We interpret \emph{events} as indicators of inter-character relationship states - for each event, we extract the latent predicate-argument structure to identify participants. For example, in the sentence ``Harry kicked Ron", ``kicked" is the \emph{event} with Harry the \textit{actor} and Ron the \textit{experiencer}.

\subsection{Circumstance}
We take inspiration from \cite{vonnegutshapes} and use implied sentiment and emotion to quantify \emph{circumstance}, the state of fortune associated with a directed pair of characters participating in an \emph{event}. Narratives and plots span a plethora of settings - each character `reacts to' as well as `influences' their unique \emph{circumstance}; it would be beyond the scope of this paper to establish a universal scale. Instead, we focus on the relative \emph{shift of circumstance} across the narrative. 

\subsection{Relation Arcs and Character Arcs}
As the story progresses, the relationship between a pair of characters evolves. A \emph{relation arc} is a plot of the shift in circumstance of a directed pair of characters across their participatory events in a narrative. While the \emph{actor} of an event influences the circumstances that affect the \emph{experiencer}, the latter's subsequent actions are captured in other \emph{relation arcs}, allowing the effect of events to trickle through multiple connected arcs forming the plot of a narrative. We posit that a character's journey at any point in the narrative can be represented by contextually assessing the amalgamated effect of their interactions with both themselves and other characters on their circumstance; a \emph{character arc} is thus defined as a pair of quantitative aggregations of all the corresponding \emph{relation arcs} of a character as both \emph{actor} and \emph{experiencer} respectively.

Fig ~\ref{fig:introduction} shows the character arc for the protagonist of the Lord of the Rings, Frodo Baggins, capturing the shift of circumstances he experiences as both an \emph{actor} and \emph{experiencer}. The events Frodo participates in change his circumstances - a drop to the valleys (minimas) represents a deterioration of circumstances, whereas a rise to the peaks (maximas) denotes an improvement. It is evident from \textbf{C} and the arc slice between \textbf{D} and \textbf{E} that actor and experiencer arcs of a character are not always aligned, allowing for the intuitive explanation of instances when characters oppose the circumstances they find themselves in, for better or for worse.

\section{MARCUS}
\label{sec:marcus}
We present MARCUS (Modelling Arcs for Understanding Stories), an event-centric NLP pipeline\footnote{ \url{https://github.com/darthbhyrava/MARCUS}} that generates character arcs from narratives. In this section, we will elaborate upon the various components of the pipeline.

\subsection{Dataset}
We failed to retrieve meaningful character arcs in our preliminary experiments with short stories containing less than 500 events. Consequently, we choose two larger datasets for character arc analysis instead, namely The Harry Potter septology and The Lord of the Rings trilogy. For each source, we ensure that we remove page numbers, unnecessary annotations, unreadable characters, weblinks, etc. The details of each dataset are detailed in Table \ref{tab:datadetails}. 

\begin{table}
\parbox{.45\linewidth}{
\centering
\caption{Dataset Details}
\begin{tabular}{c|c|c}
    Data Source & Word Count & Event Count \\
    \hline
     Harry Potter & 1,095,940 & 93,782 \\
     Lord of the Rings & 478,329 & 28,670 \\
\end{tabular}   
\label{tab:datadetails}
}
\hfill
\parbox{.45\linewidth}{
\centering
\caption{RoBERTa Sentiment Regression Model}
\begin{tabular}{c|c}
    Metric &  Score  \\
    \hline
     Mean Squared Error & 0.01620 \\
     Mean Absolute Error & 0.09693
    \end{tabular}
\label{tab:sent_regressor}
}
\end{table}

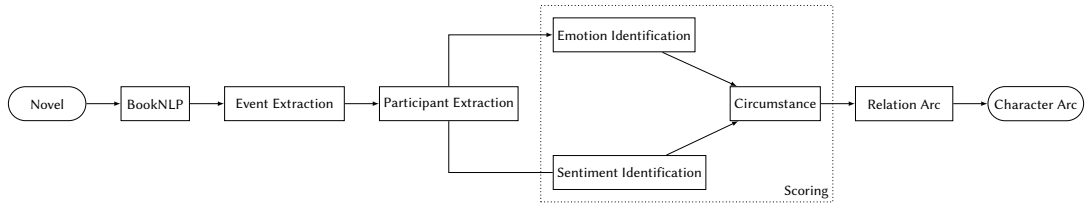
\begin{figure*}[ht]
    \centering
    \scalebox{0.45}{
    \begin{tikzpicture}[font=\large,thick]
    \node[draw,
    rounded rectangle,
    minimum width=2.5cm,
    minimum height=1cm] (block1) {Novel};

\node[draw,
    rectangle,
    right=of block1,
    minimum width=2cm,
    minimum height=1cm
] (block2) {BookNLP};

\node[draw,
    align=center,
    right=of block2,
    minimum width=3.5cm,
    minimum height=1cm
] (block3) {Event Extraction};

\node[draw,
    align=center,
    right=of block3,
    minimum width=3.5cm,
    minimum height=1cm
] (block4) {Participant Extraction};

\node[draw,
    align=center,
    below right=of block4,
    minimum width=3.5cm,
    minimum height=1cm
] (block5) {Sentiment Identification};

\node[draw,
    align=center,
    above right=of block4,
    minimum width=3.5cm,
    minimum height=1cm
] (block6) {Emotion Identification};

\node[draw,
    align=center,
    below right=of block6,
    minimum width=2.5cm,
    minimum height=1cm
] (block7) {Circumstance};

\node[draw,
    align=center,
    right=of block7,
    inner sep=10pt,
 minimum width=2.5cm,
    minimum height=1cm
] (block8) {Relation Arc};

\node[draw,
    rounded rectangle,
    align=center,
    right=of block8,
    inner sep=10pt,
    minimum width=3.5cm,
    minimum height=1cm
] (block9) {Character Arc};

\node[inner sep=10pt, draw,dotted, fit=(block5) (block6) (block7) ](scoring) {};
\node[above left] at (scoring.south east) {Scoring};

\draw[-latex] (block1) edge (block2)
    (block2) edge (block3)
    (block3) edge (block4)
    (block4) |- (block5)
    (block4) |- (block6)
    (block6) edge  (block7)
    (block5) edge (block7)
    (block7) edge (block8)
    (block8) edge (block9);
    
\end{tikzpicture}
}
    \caption{MARCUS (Modeling Arcs for Understanding Stories), an NLP pipeline that plots a character's arc as their quantitative interaction with circumstance as both actor and experiencer, represented by the proxy amalgamation of their event-centric relations across the narrative.}
    \label{fig:marcus}
\end{figure*}

\subsection{BookNLP}
BookNLP \cite{bamman-etal-2014-bayesian} is an NLP pipeline\footnote{\url{https://github.com/dbamman/book-nlp}} that scales to books and long documents (in English) and performs tasks like part-of-speech tagging (Stanford), dependency parsing (MaltParser), named entity recognition (Stanford), character name clustering (e.g., ``Tom", ``Tom Sawyer", ``Mr. Sawyer" : ``Tom Sawyer") and pronominal coreference resolution. MARCUS uses BookNLP to retrieve character occurrences and linguistic features needed for event and participant extraction. 

\subsection{Event Extraction}
\label{sec:event_id}
We feed the features processed from BookNLP into a tagger consisting of a BiLSTM model with BERT embeddings to extract events and corresponding entities from the narrative. The tagger is trained\footnote{\url{https://github.com/dbamman/litbank}} using the LitEvents dataset \cite{sims-etal-2019-literary}, which provides an annotated list of events from over 100 public domain novels from Project Gutenberg. MARCUS uses tagged events and entities for extracting participant characters, sentiment, and implied emotion.

\subsection{Participant Extraction}
We process each event with a BERT-based Semantic Role Labeller provided by AllenNLP\footnote{\url{https://docs.allennlp.org/models/main/models/structured_prediction/models/srl_bert/}} to extract the latent predicate-argument structure of the event and identify roles of participant characters. If there are multiple events in the same sentence with the same actors and experiencers, we consider only the first event to avoid redundancy. 

\subsection{Circumstance}
Every actor/experiencer pair corresponding to an event in the narrative is assigned a quantitative measure of circumstance. We argue that proxy indicators such as sentiment and implied emotion implicitly capture an event's circumstances and are specifically well suited to our focus on \emph{shift of circumstance}. Absolute measures of circumstance can therefore be interpreted as a characteristic of the genre or trope of narrative, while their shifts are a characteristic of the character's journey. Since circumstances evolve over time (or event sequences), MARCUS considers the effect of previous circumstances between characters when defining \emph{relation arcs}. 

\subsubsection{Sentiment Identification}
We pose sentiment extraction as a regression task to capture the subtleties of relationships. We fine-tune a RoBERTa model on the Stanford Sentiment Treebank (SST) \cite{socher-etal-2013-recursive} dataset to obtain a fine-grained sentiment score in the range of $0$ to $1$. The SST dataset provides $12k$ sentences and phrases with their associated sentiment scores lying between $0$ to $26$, which we normalize before training for ten epochs in a 60:20:20 split. The metrics for this model are reported in Table \ref{tab:sent_regressor}.  MARCUS uses the model to assign sentiment scores to events extracted earlier in Section \ref{sec:event_id}. 

\subsubsection{Emotion Identification}
Sentiment alone may not give us enough information about circumstance - we argue that in such cases, multi-faceted emotional states help capture shift in circumstance by leveraging the nuances of relationships. To identify emotions, we use a BERT model\footnote{\url{https://github.com/monologg/GoEmotions-pytorch}} trained on the GoEmotions \cite{Demszky2020GoEmotionsAD} dataset in a multi-label setting, as interactions can have more than one emotional undertone. The GoEmotions dataset consists of 58k Reddit comments manually annotated for 27 emotion categories: admiration, amusement, sorrow, fear, etc. MARCUS uses the confidence of the model's predicted labels as well as a manually assigned value for each label to contribute to the measure of circumstance.  These labels, in the range of $-2$ to $2$, are assigned based on intensity of emotion; higher intensity corresponds to higher absolute value. 

\subsection{Relation and Character Arcs}

MARCUS generates relation arcs by plotting the measure of circumstance, $t$, for every event, $e$, belonging to an actor/experiencer pair of characters across the narrative. 
\begin{equation}
    t_{\text{e}} = \alpha * s_{\text{e}} + \sum^{L}_{i = 1} \beta_{i} * c_{\text{i}_{\text{e}}} 
\end{equation}

where $t_{\text{e}}$ is the measure of circumstance for event $e$, $\alpha$ $\in (0, 1)$ is the sentiment co-efficient  that controls how much influence  the fine grained sentiment score should have over relation arcs, $s_{\text{e}}$ $\in (0, 1)$  is the sentiment of that event, $L$ is the total number of emotion labels for that event, $\beta_\textit{{\text{i}}}$ $\in [-2, 2]$ is the fixed score for emotion label, and $c_{\text{i}_{\text{e}}}$ $\in (0, 1)$ is the corresponding confidence score for each emotion label in the event. We run multiple experiments to choose optimal values of $\alpha$ and $\beta$: their final values are listed in the code.

We apply a window function $R(.)$ over $\textbf{t}_{\textbf{\text{e}}}$, the set of all measures of circumstance corresponding to the event set $\textbf{e}$, to calculate the relation arc, $\textbf{r}$, given by 
\begin{equation}
    \textbf{r} = R(\textbf{t}_{\text{e}}, n, p)     
\end{equation}

where $R(.)$ is the window function that helps smoothen the arcs while retaining previous state information, $n$ is the window size, and $p$ is an optional parameter for specifying order for polynomial fitting.

\begin{figure}
     \centering
     \begin{subfigure}[b]{0.45\textwidth}
         \centering
         \includegraphics[width=\textwidth]{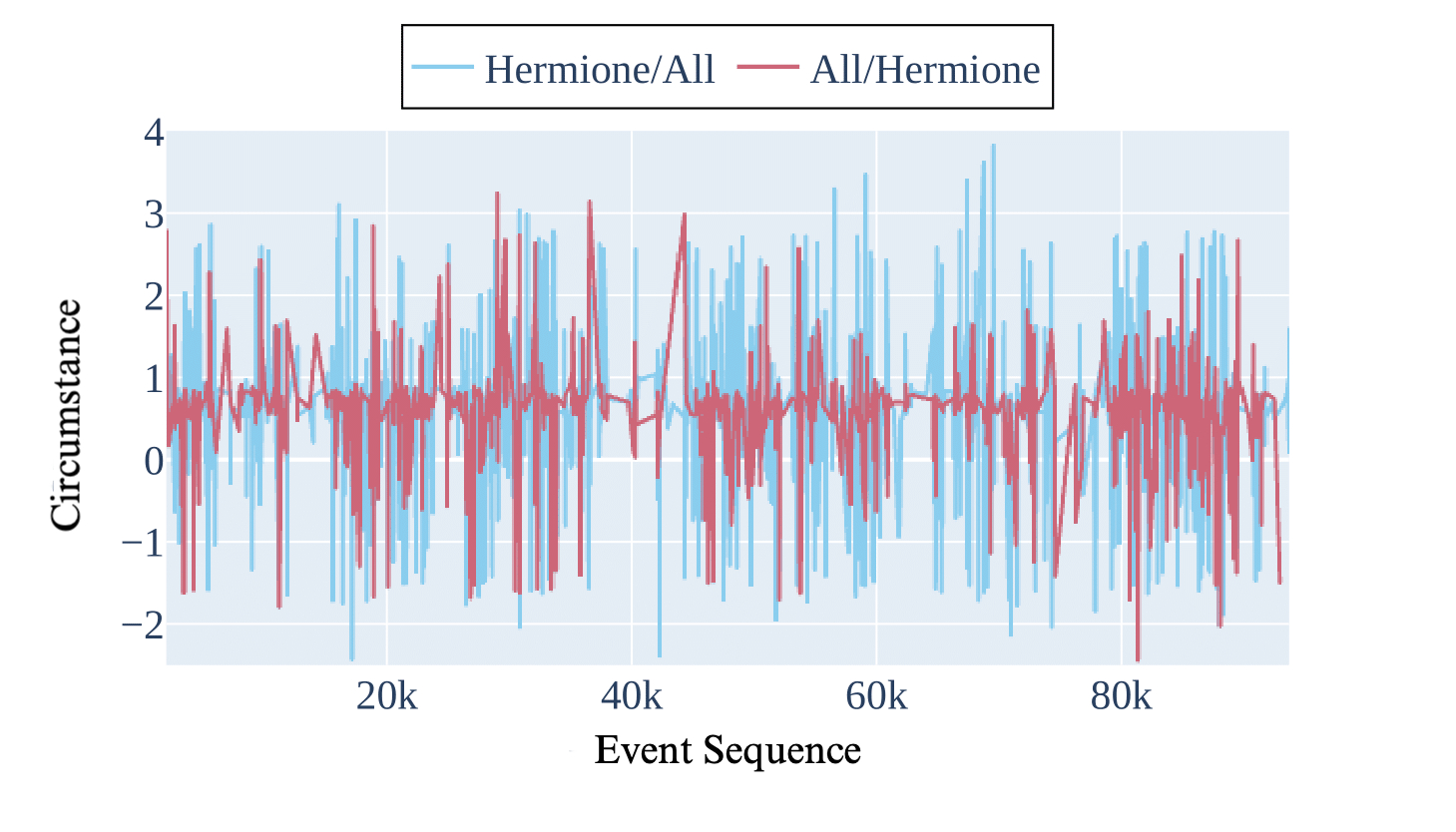}
         \caption{}
         \label{fig:hermione-vanilla}
     \end{subfigure}
     \hfill
     \begin{subfigure}[b]{0.45\textwidth}
         \centering
         \includegraphics[width=\textwidth]{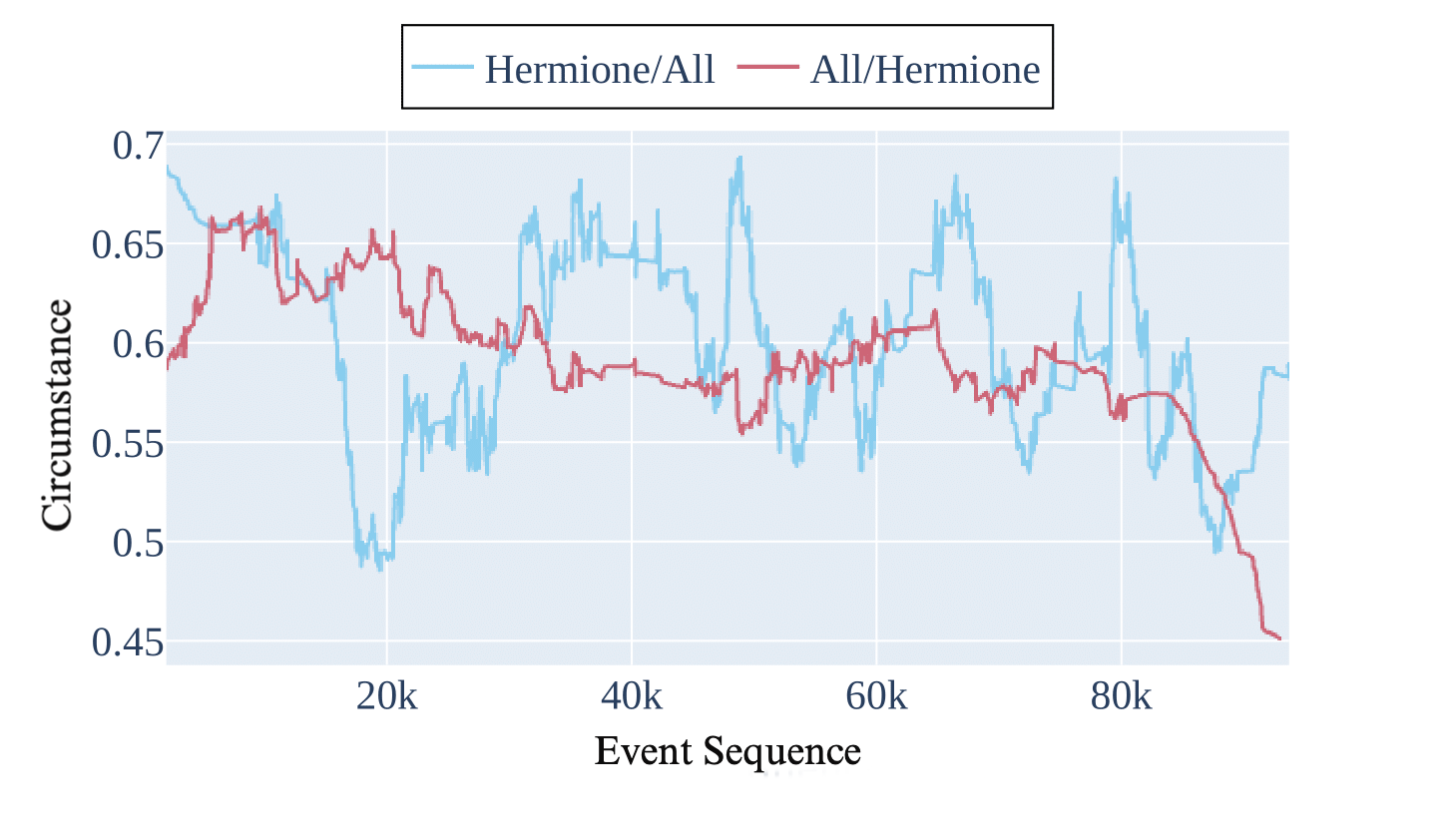}
         \caption{}
         \label{fig:hermione-savgol}
     \end{subfigure}
    \caption{(a) Character Arc for Hermione, No Rolling Function Applied; (b) Character Arc for Hermione, with a savgol filter of window size 1/10th of her event sequence length, fitted with a third degree polynomial.}
    \label{fig:hermoine}
\end{figure}

As shown in Fig \ref{fig:hermione-vanilla}, the relation arcs are too noisy without smoothing or retention of previous circumstance information. We experiment with three standard window functions: Rolling Mean, Triangular Rolling Mean and Savitzky-Golay Filter \cite{savitzky1964smoothing}. We find that the Savitzky-Golay Filter represents the narrative most accurately as seen in Fig \ref{fig:hermione-savgol}, and use the same for all arcs represented in the paper. We generate the character arc, $\textbf{c}$, by adding up the corresponding relation arcs $\textbf{r}$ of that character over all events $\textbf{e}$ in the role of actor and experiencer respectively. 
\section{Evaluation}
\label{sec:eval}
\subsection{Survey}
We ask a set of 16 human volunteers (avid fiction readers aged 20-31) to peruse both the Harry Potter and Lord of the Rings series, following which they evaluate our system by answering surveys on two tasks: \textbf{idchar}, where the volunteers are given a list of relation arcs and character pairs and asked to match the arcs to their corresponding pairs, and \textbf{idplot}, where the volunteers are given character arcs, pertinent plot events and asked to identify the points in the arc that they think represent the corresponding events.

For both these tasks mentioned above, we calculate accuracy. Task \textbf{idchar} achieved an accuracy of $71.5\%$ and task \textbf{idplot} had an accuracy of $72.9\%$. We also evaluate with the Fleiss Kappa metric where category 1 indicates a complete match, and 0 indicates otherwise. For both the tasks, we have a score of $0.675$ and $0.528$ indicating strong and moderate inter-annotator agreement, respectively. Thus, most of the volunteers consistently identified both the character pairs and the relevant points in the graph given the plot sequence.  

\subsection{Gold Labels}
We have a volunteer extremely familiar with the story annotate the first 300 events of Lord of the Rings trilogy involving Frodo Baggins as a participant character. The annotator marks each event with three labels denoting a \emph{positive}, \emph{neutral} or \emph{negative} shift in circumstance. Tables \ref{tab:pos_shifts} and \ref{tab:neg_shifts} illustrate the positive and negative shifts as tagged by our system and the corresponding gold labels for the same events provided by our annotator. Our system tends to assign positive labels to neutral events and has higher accuracy for negative shifts of circumstance.

\begin{table}
\parbox{.425\linewidth}{
\centering
\captionsetup{justification=centering}
\caption{Positive Shifts}
\begin{tabular}{c|c}
    Gold Label & Percentage \\
    \hline
     Positive & 0.36 \\
     Neutral & 0.30 \\
     Negative & 0.33 \\
\end{tabular}
\label{tab:pos_shifts}
}
\hfill
\parbox{.425\linewidth}{
\centering
\captionsetup{justification=centering}
\caption{Negative Shifts}
\begin{tabular}{c|c}
    Gold Label & Percentage \\
    \hline
     Positive & 0.12 \\
     Neutral & 0.15 \\
     Negative & 0.73 \\
\end{tabular}
\label{tab:neg_shifts}
}
\end{table}

\section{Challenges and Future Work}
We identify five notable challenges in our approach that can be addressed in future work. Firstly, since MARCUS is a sequential pipeline, it is challenging to determine the effect of errors cascading through the system quantitatively. Secondly, our rolling window makes the arcs dependent on the availability of data; event paucity in short stories or characters with low interactions hinders accurate arc generation. Thirdly, we observe in our arcs that our fine-grained events do not represent an abstract view of the discourse - a more contextual representation of events is needed. Fourth, in our understanding of a character's circumstance, localized interactions with other event-specific characters heavily influence shifts; we need an effective means of capturing the latent relative importance of character-specific interactions. And lastly, our approach does not aim to handle non-linear narratives where events are not sequentially presented.

\section{Applications}
\label{sec:conclusion}

Providing tangibility to the theoretical concept of character arcs, MARCUS can be employed in a variety of applications. Character arcs can be used in a more nuanced approach for detecting similarity between narratives by focusing on character journeys, leading to a possible improvement in book recommendations and movie recommendations based on stories and scripts. Character arcs can also help with digital enrichment in e-readers, adding to the rich metadata provided by devices like Kindle. Character arcs can also function as guidance for natural language generation tasks in the field of fiction. And lastly, they can help narrative studies by identifying character tropes and for identification of personality traits. 

\section{Conclusion}
\label{sec:conclusion}
We propose MARCUS (Modeling Arcs for Understanding Stories), an NLP pipeline that addresses the novel task of generating character arcs from narratives. We explain key concepts like \emph{events} and \emph{circumstance} and delve into the details of our event-centric approach which leverages proxy markers like sentiment and emotion. We then evaluate our pipeline, discuss challenges, elucidate future work and outline potential applications.

\bibliography{MAIN}
\end{document}